\documentclass[preprint,sigconf]{acmart}
\RequirePackage{latexrelease}
\usepackage{booktabs} 
\usepackage{fixltx2e}
\usepackage{siunitx,etoolbox}
\usepackage{multirow}
\usepackage{algorithm}
\usepackage{algorithmic}

\renewcommand\footnotetextcopyrightpermission[1]{} 
\setcopyright{none}

\begin{document}
\pagenumbering{gobble}

\title{Applying Partial-ACO to Large-scale Vehicle Fleet Optimisation}

\author{Darren M. Chitty, Elizabeth Wanner, Rakhi Parmar, Peter R. Lewis}
\affiliation{%
  \institution{Aston Lab for Intelligent Collectives Engineering (ALICE)\\Aston University, Birmingham. B4 7ET UK}
} \email{{d.chitty, e.wanner, r.parmar8, p.lewis}@aston.ac.uk}

\renewcommand{\shortauthors}{Darren M Chitty et al.}

\begin{abstract}
Optimisation of fleets of commercial vehicles with regards
scheduling tasks from various locations to vehicles can result in
considerably lower fleet traversal times. This has significant
benefits including reduced expenses for the company and more
importantly, a reduction in the degree of road use and hence
vehicular emissions. Exact optimisation methods fail to scale to
real commercial problem instances, thus meta-heuristics are more
suitable. Ant Colony Optimisation (ACO) generally provides good
solutions on small-to-medium problem sizes. However, commercial
fleet optimisation problems are typically large and complex, in
which ACO fails to scale well. \emph{Partial-ACO} is a new ACO
variant designed to scale to larger problem instances. Therefore
this paper investigates the application of \emph{Partial-ACO} on
the problem of fleet optimisation, demonstrating the capacity of
\emph{Partial-ACO} to successfully scale to larger problems.
Indeed, for real-world fleet optimisation problems supplied by a
Birmingham based company with up to 298 jobs and 32 vehicles,
\emph{Partial-ACO} can improve upon their fleet traversal times by
over 44\%. Moreover, \emph{Partial-ACO} demonstrates its ability
to scale with considerably improved results over standard ACO and
competitive results against a Genetic Algorithm.
\end{abstract}


\maketitle

\vspace{-0.25cm}
\section{Introduction}
Fleet optimisation is a common problem faced by many organisations
from delivery companies to maintenance firms such as plumbers or
builders to the medical profession performing care in the
community. The companies in question have many tasks to perform
over a geographical area and a set of vehicles with which to carry
them out. The problem is two fold, which tasks to assign to each
vehicle to carry out and the order by which each vehicle conducts
these tasks such that the traversal time of the fleet of vehicles
as a whole is minimised. Moreover, there are often challenging
constraints such as vehicular capacities, capabilities and time
windows within which tasks must be carried out.

The primary benefit from fleet optimisation is reduced costs both
in terms of fuel and labour costs with regards driving.
Optimisation can also facilitate a greater number jobs to be
undertaken by the vehicle fleet. However, an added important
benefit from reducing vehicular fleet traversal time is a
reduction in the degree of traffic on the road network of a city.
This will assist in a reduction in vehicle emission levels
experienced within urban conurbations. Indeed, the World Health
Organisation (WHO) reports the level of particulates such as
nitrogen oxides (\texorpdfstring{NO\textsubscript{x}})) in the air
of major cities are increasing markedly\footnote{Air pollution
levels rising in many of the worlds poorest
cities.\\\url{http://www.who.
int/mediacentre/news/releases/2016/air-pollution-rising/en}}.
These can cause breathing problems and have been linked to
increased rates of cardiovascular disease
\cite{requia:2018,calderon:2016}. In fact, many cities have limits
on permissible pollution, such as London which must maintain a
level of particulates below a threshold. Consequently, clean air
policies are now being pursued, such as that being considered by
Birmingham City Council\footnote{A Clean Air Zone for Birmingham
\url{https://www.birmingham.gov.uk/caz}}, and fleet optimisation
can assist in this goal.

However, optimising fleets of vehicles with constraints is NP-hard
in nature and consequently, techniques such as heuristics applied
to the problem fail to scale well with problem size. The work
presented in this paper will profile a recent advance in the Ant
Colony Optimisation (ACO) meta-heuristic technique known as
\emph{Partial-ACO} \cite{chitty:2017} and demonstrate its ability
to scale better than ACO when applied to real-world fleet
optimisation problems.

The paper is laid out as follows: Section 2 describes the fleet
optimisation problem and prior approaches for solving. Section 3
describes in detail the \emph{Partial-ACO} approach. The results
from applying \emph{Partial-ACO} to a real-world fleet
optimisation problem with steadily increasing complexity will be
demonstrated in Section 4 alongside competing meta-heuristic
methods. Finally, Section 5 sums up the benefits of
\emph{Partial-ACO} for fleet optimisation.

\section{Problem Definition and Prior Art}
Optimising vehicles through a road network has been well studied.
Initial work involved routing a single vehicle visiting a set of
differing locations once only with the minimal distance traversed
such as would be done by a salesman. Hence this type of task
became known as the Travelling Salesman Problem (TSP). An
extension of this problem is to consider multiple vehicles being
used to visit each location known as the Vehicle Routing Problem
(VRP), applicable to many real-world logistic scenarios such as
parcel deliveries and the provision of maintenance services.
Furthermore, an extension to this problem is to consider a
capacity limit to vehicles such as packages the vehicle can hold
or the amount of fuel or a working time directive. This is known
as the Capacitated Vehicle Routing Problem (CVRP). Now the problem
becomes minimising the distance travelled by all vehicles whilst
not violating the constraints on capacity. Extending further is
the consideration of multiple locations from which the vehicles
operate from, essentially depots. Vehicles can return to any of
these depots or have to return to the depot where hence they start
from. This version of the VRP is known as the Multi Depot Vehicle
Routing Problem (MDVRP) and was formulated in 1959 by Dantzig and
Ramser \cite{dantzig:1959}. The task is to assign customers to
vehicles operating from the available depots such that all
customers are serviced whilst the minimum distance is travelled by
the fleet of vehicles.

A final aspect to consider with the VRP is time windows (VRPTW)
whereby now the locations to visit must be done so within a given
start and end time. For instance a customer to visit may only be
available within a given time window. Consequently, the problem
now becomes minimising the travel time for a fleet of vehicles
whilst not violating the constraints imposed by time windows for
delivery or servicing and the capacity of the vehicle.


The MDVRP can be formally defined as a complete graph $G=(V,E)$,
whereby $V$ is the vertex set and $E$ is the set of all edges
between vertices in $V$. The vertex set $V$ is further partitioned
into two sets , $V_{c}={V_{1},...,V_{n}}$ representing customers
and $V_{d}={V_{n+1},...V_{n+p}}$ representing depots whereby $n$
is the number of customers and $p$ is the number of depots.
Furthermore, each customer $v_{i} \in V_{c}$ has a service time
associated with it and each vehicle $v_{i} \in V_{d}$ has a fixed
capacity associated with it defining the ability to fulfill
customer service. Each edge in the set $E$ has an associated cost
of traversing it represented by the matrix $c_{ij}$. The problem
is essentially to find the set of vehicle routes such that each
customer is serviced once only, each vehicle starts and finishes
from the same depot, each vehicle does not exceed its capacity to
service customers and the overall cost of the combined routes is
minimised.

\vspace{-0.25cm}
\subsection{Methods Applied to Solving the MDVRP}
The MDVRP is recognised as an NP-hard problem and hence difficult
to solve using exact mathematical approaches. Indeed, it has been
shown that for symmetric cases and a single vehicle, exact methods
do not scale beyond 50 customers and only up to a few hundred for
asymmetric cases \cite{laporte:1992}. However, some limited work
exists in the literature. Laporte \emph{et al.} formulated a
symmetric version as an integer linear program involving three
constraints and solved using branch and bound methods
\cite{laporte:1984}. Laporte \emph{et al.} also investigated
asymmetric MDVRP problems by translating to constraint assignment
problems and again solved using branch and bound
\cite{laporte:1988}. For MDVRP instances with a heterogenous fleet
of vehicles available, Dondo \emph{et al.} used mixed-integer
programming to solve this problem \cite{dondo:2003} later
extending to multiple pickups and deliveries \cite{dondo:2008} and
again to using time windows \cite{dondo:2009}. Branch and bound
methods have also been recently applied to solving MDVRP problems
with some success \cite{benavent:2013,braekers:2014}.

However, these exact methods can only solve problems with fewer
than 50 customers for symmetric instances. Consequently, heuristic
methods are more likely to be applied to larger problems.
Heuristic methods are rule of thumb approaches not guaranteed to
find optimal solutions to problems but can find solutions close to
the optimal in significantly less computational time than exact
methods. One the earliest methods uses the Clarke and Wright
savings criterion \cite{clarke:1964} whereby customers are
assigned to their nearest deport and routes between depots and
customers created which are then gradually merged into larger
routes using this savings criterion \cite{tillman:1969}. Tillman
and Hering extended this further to consider looking ahead to
consider the effect of assignments on future assignment decisions
\cite{tillman:1971}.

An alternative heuristic is a sweep method employed by Wren and
Holliday whereby each customer is assigned to their nearest depot
and the polar angle to this depot calculated \cite{wren:1972}.
Customers are sorted in ascending order of angle and iteratively
assigned to routes with least additional distance. Gillette and
Johnson used a similar approach but instead assign customers to
depots to form compact clusters \cite{gillett:1976}. Each depot
takes a turn to be an attracting centre whereby customers are
possibly reassigned when lying between two depots. Golden \emph{et
al.} superimposed a grid over the problem and then only join
vertices in adjoining cells \cite{golden:1977}. The authors used a
second approach whereby first customers are assigned to depots and
then the VRP solved for each individual depot followed by an
improvement phase. Tests were performed on problems of up to 256
customers with reasonable results. Raft used an approach whereby
the best number of vehicles to use is calculated \cite{raft:1982}.
Once this has been estimated the customers are clustered into a
matching number of groups which are each assigned to a vehicle.
2-opt local search is then used to find the optimal cluster
routes. Chao \emph{et al.} assigned customers to their closest
depot and then solve the VRPs for each individual depot using a
modified savings algorithm \cite{chao:1993}. Salhi and Sari
presented a ``multi-level composite'' heuristic which could
identify solutions similar to those found in the literature at a
fraction of the computational cost \cite{salhi:1997}. Salhi and
Nagy later use an insertion heuristic to minimise the routing cost
\cite{salhi:1999}.

\vspace{-0.25cm}
\subsection{Meta-Heuristic Approaches}
\label{sec:metaheuristic} An alternative to solving MDVRP problems
is to use meta-heuristics, which are essentially a search-based
heuristic and are problem independent. The first use of a
meta-heuristic approach to the MDVRP was implemented by Gendreau
\emph{et al.} involving tabu search solution exploring the search
space by potentially moving to a neighbouring solution even with
degradation in the quality \cite{gendreau:1994}. Solutions
recently examined are declared \emph{tabu} for a number of
iterations such that cycling is avoided. Renaud \emph{et al.} also
used tabu search with the additional constraint that a vehicle
route cannot exceed a certain amount to adhere to working time
directives \cite{renaud:1996}.

The most commonly used meta-heuristic approach to solving VRP-type
problems is a Genetic Algorithm (GA) \cite{Holland:1975}. A
population based approach which uses the principles of Darwinian
evolution such as natural selection, crossover and mutation to
successively improve the quality of a population of solutions. The
first use of a GA to solve the MDVRP was proposed by Filipec
\emph{et al.} for non-fixed destination instances
\cite{filipec:1997}. Salhi et al. also used a GA to solve MDVRP
instances \cite{salhi:1998}. Skok \emph{et al.} used a GA and
compared six differing crossover operators \cite{skok:2000} later
applying the approach to a real-world problem with 248 customers
and three depots \cite{skok:2001}. An alternative combining
clustering with a GA was proposed by Thangiah and Salhi whereby
the GA defines clusters of customers and then routes were found by
solving the resulting TSP \cite{thangiah:2001}.

Ho \emph{et al.} proposed a hybrid approach whereby the initial
population of solutions is generated using the Clarke and Wright
saving algorithm with the nearest neighbour heuristic proving
better than random generation \cite{ho:2008}. Surekha and Sumathi
used a similar implementation obtaining comparable results to the
literature using a set of classical problems \cite{surekha:2011}.
Alba and Dorronsoro used a cellular GA approach with the
population organised into a grid and genetic operations only occur
within small neighbourhoods successfully improving upon the best
known solutions for nine classical instances of the MDVRP
\cite{alba:2006}. A wide survey of the use of GAs for solving
MDVRPs can be found in Karakati{\v{c}} and Podgorelec
\cite{karakativc:2015}.

The largest competing meta-heuristic approach to a GA applied to
the MDVRP is based on the foraging behaviour of ants known as Ant
Colony Optimisation (ACO) \cite{dorigo:1997}. Essentially, ants
operate as a collective in finding food by depositing a pheromone
as they move. This pheromone builds up as ants successively
traverse the same area helping guide ants to promising areas
containing food. Yalian implemented a version of ACO which
combined a scanning algorithm to initially assign jobs to nearest
depots and then combined with GA inspired mutation operators and
local search \cite{yalian:2016}. Results demonstrated an
improvement over standard ACO for six problems. Yu \emph{et al.}
also achieved better results from a parallel implementation of an
improved ACO algorithm whereby a virtual depot is used with
mutation operators applied to problems with up to 360 customers
and 6 depots \cite{yu:2011}. Yao \emph{et al.} also used a single
depot approach to solve a real-world MDVRP finding multiple routes
from this central depot and then assigning the individual routes
to depots \cite{yao:2014}. Calvete \emph{et al.} contrasted two
ACO approaches, one using a single \emph{super depot} and the
other a GA-ACO approach whereby customers are assigned to depots
using a GA and then the routing solving by ACO for problems of up
to 288 customers and 6 depots \cite{calvete:2011}. The GA-ACO
proved the better algorithm for solving the MDVRP. Stodola uses
ACO augmented with an additional route optimisation step to solve
a modification of the MDVRP whereby the goal is to minimise the
longest route experienced by a vehicle \cite{stodola:2018}.

A final meta-heuristic of note for solving the MDVRP is based on
the behaviours of swarms of insects and flocks of birds known as
Particle Swarm Optimisation (PSO) \cite{eberhart:1995}. This
algorithm searches the fitness landscape using a collection of
particles that fly around the landscape with each particle
consisting of a position and velocity both described as a set of
multi-dimensional coordinates. Each particle's position is updated
according to its velocity and then the velocity is updated
according to a particles best found solution and the globally best
found. Surekha and Sumathi applied PSO to the MDVRP grouping
customers to nearest depots and initially routing using the Clark
and Wright saving method and these routes were then improved using
PSO \cite{surekha:2011}. Wenjing and Ye achieved better results
with improved inertia weights and mutation operators
\cite{wenjing:2010}. Geethra \emph{et al.} apply a modified PSO
algorithm whereby initial particles are generated using k-means
and nearest neighbour heuristics \cite{geetha:2012} improving
further with a nested PSO approach whereby a master swarm assigns
customers to groups serviced by a depot and slave swarms then find
the optimal route for each group applied to problems with 160
customers and 4 depots \cite{geetha:2013}.

\section{The Partial-ACO Approach}
\emph{Partial-ACO} is a derivative of the Ant Colony Optimisation
(ACO) technique. ACO will be profiled first followed by a
description of the \emph{Partial-ACO} approach.

\subsection{Ant Colony Optimisation}
The predominate alternative meta-heuristic approach to using a GA
to solve fleet optimisation/MDVRP problems is to consider an
algorithm based on the study of how ants forage for food known as
Ant Colony Optimisation (ACO) \cite{dorigo:1997}. Ants deposit
pheromone on the paths they take which enables them to
successfully find food and bring it back to their nest. Pheromone
levels will build up on edges leading to food enabling further
ants from the colony to discover the food source. The ACO
algorithm applied to fleet optimisation and MDVRP type problems
involves simulated ants moving through a graph $G$
probabilistically visiting every customer once only and depositing
pheromone as they move. The level of pheromone an ant deposits on
the edges $E$ of graph $G$ is defined by the quality of the
solution the given ant has generated. Ants probabilistically
decide which location or customer to visit next using this
pheromone level on the edges of graph $G$ plus heuristic
information based upon the distance between an ant's current
position and customers left to visit. An \emph{evaporation} effect
is used to prevent pheromone levels building up too much reaching
a state of local optima. Therefore, the ACO algorithm consists of
two stages, the first \emph{solution construction} and the second
stage \emph{pheromone update}.

The solution construction stage involves $m$ ants constructing
complete solutions to the fleet optimisation / MDVRP type
problems. Ants start from depot of a random vehicle from the fleet
and iteratively make probabilistic choices using the \emph{random
proportional rule} as to which customer to visit next or to return
to the depot whereby the probability of ant $k$ at point $i$
visiting point $j\in N^{k}$ is defined as:

\begin{eqnarray}
    p_{ij}^{k}=\frac{[\tau_{ij}]^{\alpha}[\eta_{ij}]^{\beta}}{\sum_{l\in
N^{k}}[\tau_{il}]^{\alpha}[\eta_{il}]^{\beta}}
\end{eqnarray}
where $[\tau_{il}]$ is the pheromone level deposited on the edge
leading from location $i$ to location $l$; $[\eta_{il}]$ is the
heuristic information consisting of the distance between customer
or depot $i$ and customer or depot $l$ set at $1/d_{il}$; $\alpha$
and $\beta$ are tuning parameters controlling the relative
influence of the pheromone deposit $ [\tau_{il}]$ and the
heuristic information $[\eta_{il}]$.

The process is repeated for each vehicle. Once all ants have
completed the solution construction stage, pheromone levels on the
edges $E$ of graph $G$ are updated. First, evaporation of
pheromone levels upon every edge of graph $G$ occurs whereby the
level is reduced by a value $\rho$ relative to the pheromone upon
that edge:
\begin{eqnarray}
\tau_{ij}\leftarrow(1-\rho)\tau_{ij}
\end{eqnarray}

where $\rho$ is the \emph{evaporation rate} typically set between
0 and 1. Once this evaporation is completed each ant $k$ will then
deposit pheromone on the edges it has traversed based on the
quality of the solution found:
\begin{eqnarray}
\tau_{ij}\leftarrow\tau_{ij}+\sum_{k=1}^{m}\Delta \tau_{ij}^{k}
\end{eqnarray}
where the pheromone ant $k$ deposits, $\Delta \tau_{ij}^{k}$ is
defined by:
\begin{eqnarray}
\Delta \tau_{ij}^{k}&=&\left\{
\begin{array}{ll}
1/C^{k}, & \mbox{if edge $(i,j)$ belongs to $T^{k}$} \\
0, & \mbox{otherwise} \\
\end{array}
\right.
\end{eqnarray}
where $1/C^{k}$ is the quality of ant $k$'s solution $T^{k}$. This
methodology ensures that better solutions found by an ant result
in greater levels of pheromone being deposited on those edges.


\subsection{Partial-ACO}
Although ACO has been a highly successful meta-heuristic approach
there are two main issues with the technique. The first is the
requirement for a pheromone matrix to be held in memory of size
$n^{2}$. Hence, if $n$ becomes extremely large there will not be
enough memory in a modern computer to hold this matrix. For
example, for a 100,000 customer problem and using a float datatype
which is four bytes in memory, approximately 37GB of memory will
be required to hold this matrix.

A second scalability factor with ACO is the length of the required
solutions as the size of the problem increases. At each step an
ant needs to probabilistically decide which unvisited location to
visit next by collating the pheromone on all unvisited edges in
graph $G$. The ant then moves to the next location using a random
value combined with the pheromone levels. Thus it can be
postulated that an ant will probabilistically make a poor decision
over which location to visit next at some point when constructing
a solution even with high concentrations of pheromone on the
correct edge. However, even if the occurrence of a poor decision
being made by ant at a decision point is of a low probability, as
the problem sizes increases the greater number of decisions an ant
needs to make and hence the chance an ant will make a poor choice.
But to construct an optimal solution every one of an ant's
decisions will need to be the optimal choice. Therefore, it can be
hypothesized that ACO by its probabilistic nature will be less
likely to find the optimal solution the larger in size the
required solution becomes.

Moreover, the computational cost of probabilistically making a
decision at each step of constructing a solution increases
quadratically. At each step an ant needs to compare the pheromone
levels and a random probability for every unvisited edge. Thus for
a five customer TSP instance, at the first step four comparisons
are needed with the ant selecting the best edge. At the next step
three comparisons are required and so forth. In total, nine
comparisons are required to construct a complete solution. This is
similar to the triangular number sequence described as
$(n(n-1)/2)$. Thus, for a 100,000 customer problem, nearly five
billion pheromone edge comparisons will be required to construct a
complete solution. Indeed, the original author of ACO noted this
computational complexity proposing a variant known as Ant Colony
System (ACS) \cite{dorigo:1997} whereby the neighbourhood of
unvisited cities is restricted. A \emph{candidate list} approach
is used whereby at each decision point made by an ant, only the
closest cities are considered. If these have already been visited
then normal ACO used. This approach significantly reduces the
computational complexity but is reliant on accurate heuristic
information to define the neighbourhood.

To address these issues a derivative of ACO known as
\emph{Partial-ACO} has been proposed which enables ACO to be
successfully applied to TSP instances of up to 200,000 cities
\cite{chitty:2017}. \emph{Partial-ACO} operates in a similar
manner to ACO but does not use a pheromone matrix addressing the
first scalability problem with ACO. Instead, pheromone is
calculated from a population of ants and their respective
solutions and their associated quality. This is effectively
population based ACO (P-ACO) \cite{guntsch:2002} whereby a
population of ant solutions is maintained in a First In First Out
manner. If a new best solution is found then it is inserted into
the population and the oldest one removed. However,
\emph{Partial-ACO}  maintains the population of ant solutions
differently in that each ant has a \emph{local memory}
($l_{best}$) of the best solution it has personally found during
the search process operating essentially as a \emph{steady state}
process. This is similar in effect to Particle Swarm Optimisation
(PSO) \cite{eberhart:1995} whereby particles use both their local
best solution and a global best to update their position.

Pheromone deposit also operates differently to ACO as pheromone
cannot \emph{build up} on the edges of graph $G$. To account for
this issue, pheromone deposit of ant $k$ on an edge $E$ of graph
$G$ is related to the quality of the solution $l_{best}^{k}$
related to the best solution found so far, $g_{best}$. Hence, the
amount of pheromone ant $k$ deposits, $\Delta \tau_{ij}^{k}$ is
defined by:\vspace{-0.25cm}

\begin{eqnarray}
\Delta \tau_{ij}^{k}&=&\left\{
\begin{array}{ll}
(g_{best}/l_{best}^{k})^{\alpha}, & \mbox{if edge $(i,j)$ belongs to $T^{k}$} \\
0, & \mbox{otherwise} \\
\end{array}
\right.
\end{eqnarray}
where $(g_{best}/l_{best}^{k})$ is the quality of ant $k$'s
locally best found solution in relation to the globally best found
solution and $\alpha$ is a parameter controlling the influence of
pheromone. This mechanism ensures that pheromone level deposits do
not alter as improved solutions are found and also do not suffer
from any scalability issues.

Since there is no pheromone matrix, an ant $k$ when building a
solution has to reconstruct the pheromone levels on the edges from
its current location to all unvisited locations. This involves
iterating through all the $l_{best}$ solutions of the ant
population and finding the edges taken from the current location
and depositing the pheromone if that location has not yet been
visited. Moreover, for each location in an ant's solution there is
an edge taken from that given location but there is also an edge
taken to arrive at the given location. Consequently, these edges
are also taken into consideration when constructing the pheromone
levels on edges.

\begin{figure}[!ht]
\centering
\includegraphics[width=0.30\textwidth]{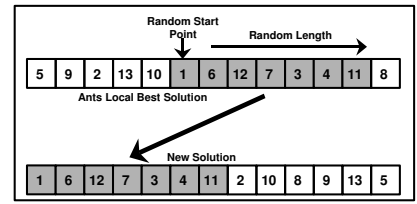}
\scriptsize \caption{An illustration of the \emph{Partial-ACO}
methodology.} \label{fig:PartialACO}
\end{figure}\vspace{-0.25cm}

The second component of \emph{Partial-ACO} addresses the error
probability during solution construction and the computational
complexity of ACO. Recall that for a hundred thousand job problem,
the number of decisions that need to be made are of an equal
number whereby there is a probability of making a poor decision.
Secondly, the number of pheromone comparisons required to build a
complete solution for a problem consisting of a hundred thousand
jobs requires nearly five billion comparisons. Consequently,
\emph{Partial-ACO} proposes that an ant when constructing a new
solution takes its current $l_{best}$ tour and retains part of
this tour and then completes the rest in the same probabilistic
manner as described earlier. This is to some degree similar to
crossover in a GA whereby a complete section of a parent solution
is copied into the child solution. To achieve this \emph{partial}
reconstruction of a solution firstly a random point is selected in
the $l_{best}$ tour. Secondly, a random length of the tour to be
retained is decided and this section is copied into the new
solution. The remaining part of the solution is then constructed
as normal. This process is illustrated in Figure
\ref{fig:PartialACO}.

If this partially reconstructed solution has greater quality then
the current $l_{best}$ for the given ant then this new solution
replaces the ant's $l_{best}$ solution. To highlight the
computational advantage of this technique, consider retaining 50\%
of solutions for a 100,000 job problem. In this instance only
50,000 probabilistic decisions now need to be made and only 1.25
billion pheromone comparisons would be required, a reduction of
75\%. In fact, the potential computational cost savings are
quadratic in nature. An overview of the \emph{Partial-ACO}
technique is described in Algorithm \ref{alg:PartialACO}.

\begin{algorithm}[]
\caption{\emph{Partial-ACO}} \label{alg:PartialACO}
\begin{algorithmic}[1]
    \FOR{each ant}
        \STATE{Generate an initial solution probabilistically}
    \ENDFOR
   \FOR{number of iterations}
        \FOR{each ant $k$}
            \STATE{Select uniform random start point from $l_{best}^{k}$ solution}
        \STATE{Select uniform random length of $l_{best}^{k}$ to preserve}
        \STATE{Copy $l_{best}^{k}$ points from start for specified length}
        \STATE{Complete remaining aspect probabilistically}
        \STATE{If new solution better than $l_{best}^{k}$ then update $l_{best}^{k}$}
    \ENDFOR
\ENDFOR \STATE{Output best $l_{best}$ solution (the $g_{best}$
solution)}
\end{algorithmic}
\end{algorithm}

\subsection{Partial-ACO Applied to Fleet Optimisation}

\begin{figure}[!ht]
\centering
\includegraphics[width=0.30\textwidth]{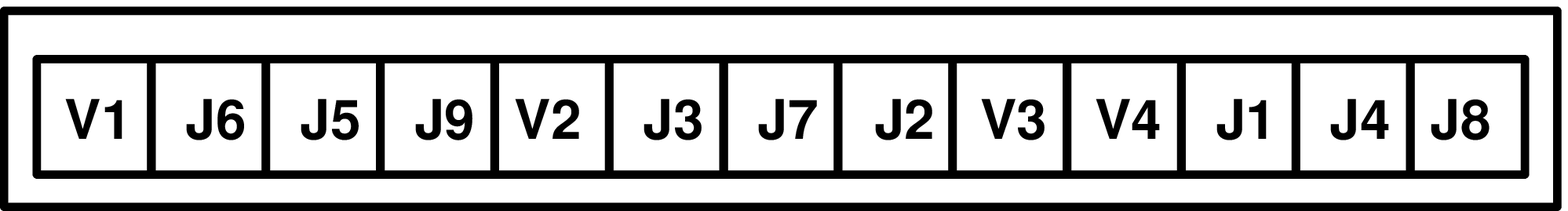}
\scriptsize \caption{Example solution representation.}
\label{fig:Representation}
\end{figure}

Fleet optimisation is in effect the MDVRPTW problem whereby there
are a range of vehicles operating from a number of depots that
need to be assigned jobs to complete such that all jobs are
completed within their time windows and the total traversal time
of the fleet of vehicles is minimised. To represent this problem a
solution will consist of a set of vehicles and the list of jobs to
be completed in the order they need to be completed. This
representation is depicted in Figure \ref{fig:Representation}
whereby \emph{V} relates to a vehicle and \emph{J} relates to a
job. It can be observed that the first vehicle in the fleet will
undertake jobs 6, 5 and 9, the second vehicle jobs 3, 7, and 2 and
so forth. Note that vehicle 3 is not assigned any jobs.

An ant will build a solution in a similar manner to a TSP instance
in that the number of locations to visit equates to the number of
jobs and the number of vehicles. Each of these locations must be
visited once only. To start a solution a vehicle must be selected
first. From this point ants probabilistically move to unfulfilled
jobs or another vehicle at which point the current vehicle returns
to its depot and a new sequence of jobs to be assigned to this new
vehicle commences.

Once a new solution has been generated by \emph{Partial-ACO} its
quality needs to be assessed. The quality is measured using two
objectives, the first of which is to maximise the number of jobs
correctly performed within their given time window. Any missed
jobs are assigned a penalty. The second objective is the total
traversal time of the fleet of vehicles. The goal is to minimise
both the number of missed jobs and the traversal time of the
vehicular fleet. Reducing the number of missed jobs is considered
the primary objective.

Pheromone deposit is calculated using the two objectives to be
optimised, the number of customers serviced and the length of the
traversal time between customers. A penalty based function will be
utilised for the first objective whereby any customers that have
not been serviced due to capacity limitations or missing the time
window will be penalised by the predicted job time. The secondary
objective objective is to minimise the time the fleet of vehicles
spend traversing the road network between jobs. Therefore, the
quality of a tour can be described as:

\begin{eqnarray}
C^{k}=(S-s^{k}+1)*L^{k}
\end{eqnarray}
where $S$ is the total amount of time of jobs to be serviced,
$s^{k}$ is the amount of job service time achieved by ant $k$'s
solution and $L^{k}$ is the total traversal time of the fleet of
vehicles of ant $k$'s solution. Clearly, if ant $k$ has achieved
the primary objective of fulfilling all customer demand $S$ then
$C^{k}$ becomes merely the total traversal time of the fleet of
vehicles which needs to be minimised.

\begin{table}[!ht]
\footnotesize \centering \caption{Real-world problem scenarios as
supplied by a Birmingham maintenance company. Scenarios described
in terms of the vehicles available, the number of customers to
service, the total predicted service time required and the total
travel time using the company's current
scheduling.\vspace{-0.25cm}} \label{tab:ProblemData}
\begin{tabular}{c
                c
                c
                c
                c}
\toprule
{Problem} & {Number} & {Number} & {Total Job} & {Total Fleet} \\
& {of} & {of} & {Servicing} & {Traversal Time}\\
 & {Vehicles} & {Jobs} & {(hh:mm)} & {(hh:mm)} \\
 \midrule
Week\_1 & 8 & 77 & {47:09} & {31:12} \\
Week\_2 & 8 & 79 & {48:24} & {22:49} \\
Week\_3 & 8 & 81 & {48:33} & {19:54} \\
Week\_4 & 8 & 61 & {54:52} & {25:55} \\
Fortnight\_1 & 16 & 156 & {95:33} & {54:01} \\
Fortnight\_2 & 16 & 138 & {102:01} & {57:07} \\
Fortnight\_3 & 16 & 160 & {96:57} & {42:43} \\
Fortnight\_4 & 16 & 142 & {103:25} & {45:49} \\
ThreeWeek\_1 & 24 & 237 & {144:06} & {73:55} \\
ThreeWeek\_2 & 24 & 217 & {150:25} & {79:56} \\
ThreeWeek\_3 & 24 & 219 & {150:34} & {77:01} \\
ThreeWeek\_4 & 24 & 221 & {151:49} & {68:38} \\
Month\_1 & 32 & 298 & {198:58} & {99:50} \\
\bottomrule
\end{tabular}
\end{table}

\begin{table*}[!ht]
\footnotesize \centering \caption{Parameters used throughout for
the GA, ACO, \emph{Partial-ACO} and \emph{Partial-ACO$^{PH}$}
approaches unless otherwise stated.\vspace{-0.25cm}}
\begin{tabular}{p{2.75cm} p{1cm} p{2cm} p{1cm} p{2cm} p{1cm} p{2cm} p{1cm}}
\toprule
\multicolumn{2}{c}{Genetic Algorithm} & \multicolumn{2}{c}{Max Min Ant System} & \multicolumn{2}{c}{Partial-ACO} & \multicolumn{2}{c}{Partial-ACO$^{PH}$}\\
\midrule
Population Size & 192 & Number of Ants & 192 & Number of Ants & 32 & Number of Ants & 192\\
Max Generations & 1,000,000 & Max Iterations & 1,000,000 & Max Iterations & 6,000,000 & Max Iterations & 1,000,000\\
Tournament Size & 5 & $\alpha$ & 1.0 & $\alpha$ & 3.0 & $\alpha$ & 1.0\\
Crossover Probability & 0.5 & $\beta$ & 1.0 & $\beta$ & 1.0 & $\beta$ & 1.0\\
Mutation Probability & 0.5 & $\rho$ & 0.02 & & & $\rho$ & 0.5\\
\bottomrule
\end{tabular} \centering
\label{tab:params}
\end{table*}

\sisetup{ table-number-alignment=center,
separate-uncertainty=true, table-figures-integer = 2,
table-figures-decimal = 2}
\begin{table*}
\footnotesize \centering \caption{The results from each
meta-heuristic approach regards optimising the schedules of the
real-world  scenarios in terms of the service time demand achieved
and the reduction in traversal time of the vehicle fleet over the
original scheduling.\vspace{-0.25cm}}
\begin{tabular}{c
                S[separate-uncertainty,table-figures-uncertainty=1,table-figures-integer = 2, table-figures-decimal = 2, table-column-width=15mm]
                S[separate-uncertainty,table-figures-uncertainty=1,table-figures-integer = 2, table-figures-decimal = 2, table-column-width=17mm]
                S[separate-uncertainty,table-figures-uncertainty=1,table-figures-integer = 2, table-figures-decimal = 2, table-column-width=15mm]
                S[separate-uncertainty,table-figures-uncertainty=1,table-figures-integer = 2, table-figures-decimal = 2, table-column-width=17mm]
                S[separate-uncertainty,table-figures-uncertainty=1,table-figures-integer = 2, table-figures-decimal = 2, table-column-width=15mm]
                S[separate-uncertainty,table-figures-uncertainty=1,table-figures-integer = 2, table-figures-decimal = 2, table-column-width=17mm]
                S[separate-uncertainty,table-figures-uncertainty=1,table-figures-integer = 2, table-figures-decimal = 2, table-column-width=15mm]
                S[separate-uncertainty,table-figures-uncertainty=1,table-figures-integer = 2, table-figures-decimal = 2, table-column-width=17mm]}
\toprule
{\multirow{3}{1.2cm}{Problem}} & \multicolumn{2}{c}{Genetic Algorithm} & \multicolumn{2}{c}{Max Min Ant System} & \multicolumn{2}{c}{Partial-ACO} & \multicolumn{2}{c}{Partial-ACO$^{PH}$}\\
\cmidrule{2-9} & {Job Time} & {Traversal Time} & {Job Time} & {Traversal Time} & {Job Time} & {Traversal Time} & {Job Time} & {Traversal Time}\\
& {Serviced (\%)} & {Reduction (\%)} & {Serviced (\%)} & {Reduction (\%)} & {Serviced (\%)} & {Reduction (\%)} & {Serviced (\%)} & {Reduction (\%)}\\

\midrule
Week\_1 & 100.00 \pm 0.00 & 28.10 \pm 7.15 & 100.00 \pm 0.00 & 33.62 \pm 3.39 & 100.00 \pm 0.00 & 32.29 \pm 3.77 & 100.00 \pm 0.00 & 37.14 \pm 2.15 \\
Week\_2 & 100.00 \pm 0.00 & 28.06 \pm 10.60 & 100.00 \pm 0.00 & 30.7 \pm 4.85 & 100.00 \pm 0.00 & 22.39 \pm 7.84 & 100.00 \pm 0.00 & 38.20 \pm 2.88 \\
Week\_3 & 100.00 \pm 0.00 & 27.47 \pm 5.17 & 100.00 \pm 0.00 & 31.48 \pm 4.68 & 100.00 \pm 0.00 & 28.75 \pm 5.55 & 100.00 \pm 0.00 & 40.43 \pm 3.36 \\
Week\_4 & 100.00 \pm 0.00 & 19.19 \pm 6.67 & 100.00 \pm 0.00 & 26.27 \pm 4.56 & 100.00 \pm 0.00 & 24.28 \pm 6.84 & 100.00 \pm 0.00 & 12.11 \pm 10.26 \\
Fortnight\_1 & 100.00 \pm 0.00 & 25.38 \pm 5.93 & 100.00 \pm 0.00 & 23.84 \pm 7.46 & 99.98 \pm 0.10 & 23.12 \pm 6.30 & 100.00 \pm 0.00 & 36.43 \pm 1.37 \\
Fortnight\_2 & 100.00 \pm 0.00 & 27.33 \pm 4.32 & 100.00 \pm 0.00 & 22.55 \pm 5.01 & 99.98 \pm 0.09 & 25.81 \pm 4.72 & 100.00 \pm 0.00 & 30.56 \pm 1.91 \\
Fortnight\_3 & 100.00 \pm 0.00 & 26.58 \pm 5.50 & 100.00 \pm 0.00 & 28.64 \pm 4.99 & 100.00 \pm 0.00 & 27.27 \pm 5.49 & 100.00 \pm 0.00 & 36.99 \pm 2.16 \\
Fortnight\_4 & 100.00 \pm 0.00 & 24.83 \pm 4.53 & 100.00 \pm 0.00 & 25.02 \pm 4.49 & 100.00 \pm 0.00 & 29.7 \pm 6.64 & 100.00 \pm 0.00 & 36.14 \pm 1.99 \\
ThreeWeek\_1 & 100.00 \pm 0.00 & 28.94 \pm 4.01 & 99.81 \pm 0.18 & -11.43 \pm 7.62 & 100.00 \pm 0.00 & 17.13 \pm 4.47 & 99.98 \pm 0.07 & 7.37 \pm 4.94 \\
ThreeWeek\_2 & 100.00 \pm 0.00 & 26.52 \pm 4.35 & 99.89 \pm 0.20 & -4.04 \pm 3.89 & 99.80 \pm 0.20 & 16.48 \pm 5.18 & 99.95 \pm 0.10 & 8.44 \pm 8.38 \\
ThreeWeek\_3 & 100.00 \pm 0.00 & 29.62 \pm 4.21 & 99.95 \pm 0.11 & 7.33 \pm 6.56 & 99.91 \pm 0.21 & 20.64 \pm 2.36 & 100.00 \pm 0.00 & 20.01 \pm 2.71 \\
ThreeWeek\_4 & 100.00 \pm 0.00 & 26.26 \pm 4.13 & 99.86 \pm 0.18 & -2.36 \pm 5.92 & 99.84 \pm 0.27 & 18.00 \pm 7.84 & 100.00 \pm 0.00 & 13.74 \pm 3.84 \\
Month\_1 & 100.00 \pm 0.00 & 29.23 \pm 4.18 & 99.76 \pm 0.18 & -17.85 \pm 3.75 & 99.82 \pm 0.18 & 19.60 \pm 4.09 & 98.04 \pm 0.49 & -26.59 \pm 8.48 \\
\bottomrule
\end{tabular} \centering
\label{tab:InitialResults}
\end{table*}

\section{Results}
\label{sec:Results}
\subsection{Problem Description and Setup}
To evaluate the application of \emph{Partial-ACO} to fleet
optimisation a real-world problem will be considered taking data
from a maintenance company based in the Birmingham area which has
multiple vehicles and multiple customers requiring periodic
external maintenance of properties. Each vehicle starts from a
given latitude and longitude, essentially representing its depot,
and must return to this location when it has finished servicing
customers. Each customer is also defined by a location consisting
of a latitude and longitude coordinate and a job duration
predicting the length of time the job will take take and in some
cases, a time window for when the jobs must be completed. The
speed of travel of a van between maintenance jobs is defined as an
average of 13kph to account for city traffic etc. There is a also
a hard start time and end time to a given working day defined as
08:00 and 19:00 hours. Customer jobs cannot be started or finished
before or after these start and end times. However, a vehicle can
depart its depot prior to the start time in order to be able to
commence a scheduled task at 08:00 hours. Equally a vehicle can
finish a job at 19:00 and return to its depot after this time
irrespective of how long it takes to get back.

The data supplied by the company is split into a number time
periods of jobs and vehicle availability. For each problem, the
company has supplied the actual division of jobs between vehicles
used and the order that the jobs were conducted. This facilitates
a \emph{ground truth} to be established of vehicle usage for the
company and consequently real-world reductions to be ascertained
from using a meta-heuristical optimisation approach. In this
instance, the company divided jobs between vehicles in a
geographical manner which they considered correct and then derived
the route for each vehicle to complete its assigned tasks by
directing it to do the job furthest from its depot and then work
its way back only deviating from this if there are jobs with time
windows. The details of each of the individual problems supplied
by the company are shown in Table \ref{tab:ProblemData}.
Approximately one in ten of the jobs has a defined time window
whereby the job must be completed.

To evaluate the effectiveness of \emph{Partial-ACO} it will be
compared to both ACO and GA meta-heuristic approaches. In this
instance the ACO approach will be based upon the max min ant
system (MMAS) \cite{stutzle:2000} which only updates the pheromone
deposits for the best found so far solution at each iteration
rather than using the whole population. Moreover, the pheromone is
restricted to minimum and maximum levels. The GA approach will use
the crossover operators cyclic crossover (CX), order crossover
(OX) and partially mapped crossover (PMX). Mutation will consist
of several operators, swapping two solution points, reversing the
order between two points and inserting a solution point elsewhere
into the solution. The GA will operate in a \emph{steady state}
manner whereby child solutions only replace parents if their
quality is better.

Additionally, a second implementation of \emph{Partial-ACO} will
be considered which does use a pheromone matrix to be used by ants
to probabilistically make decisions. In all other respects the
approach operates in the same manner as standard
\emph{Partial-ACO} with each ant maintaining a memory of its
locally best found solution $l_{best}$ and at every iteration
these locally best solutions update the pheromone according to the
quality of their individual solutions. This approach will be
termed \emph{Partial-ACO$^{PH}$}.

The parameters governing the operation of each of the four
meta-heuristic approaches are described in Table \ref{tab:params}.
A population size or number of ants of 192 is used as each
approach is parallelised and is executed on a processor with 16
threads so the size needs to be a multiple of 16. Furthermore,
\emph{Partial-ACO} uses only 32 ants as this is similar to the
number used in the original work \cite{chitty:2017} and was found
to be highly effective. To ensure the same number of solutions are
evaluated, \emph{Partial-ACO} is executed for six times the number
of iterations as the other meta-heuristics. Experiments throughout
were conducted using an AMD Ryzen 2700 processor with each
approach using 16 parallel threads of execution in order to
utilise all the available processor cores of the processor. The
algorithms were compiled using Microsoft C++. Experiments are
averaged over 25 individual execution runs for each problem with a
differing random seed used in each instance.
\newpage
\subsection{Initial Results}
To begin with, both approaches will be tested on each of the
single day problems to test their ability to find solutions better
than those currently used by the Birmingham company that supplied
the data. These are described in terms of percentage of total
customer service demand achieved, the percentage reduction in time
spent traversing the road network for the fleet of vehicles when
compared the company's original solution. These results are shown
in Table \ref{tab:InitialResults}. From these results it can be
firstly observed that both the GA and ACO approaches have improved
upon the approach utilised by the given company to schedule jobs
to its vehicles for the smaller problem instances. However, ACO
was unable to scale to the larger problem instances as postulated
being unable to consistently find solutions which service all the
customer tasks within their time windows and additionally find
solutions with larger traversal times than the company's original
scheduling. The GA proved better at scaling by achieving better
results than ACO for the larger problem instances and being able
to service all customer tasks for all the problem instances.

With regards the \emph{Partial-ACO} results, both variants were
able to improve upon standard ACO demonstrating their ability to
scale better. However, both techniques were unable to consistently
service all customers within their time windows.
\emph{Partial-ACO$^{PH}$} proved better than the non-pheromone
matrix implementation for the smaller problem instances but proved
less capable of scaling up to the larger problem instances. The
best reduction in total fleet traversal time over the original
company scheduling is achieved by \emph{Partial-ACO$^{PH}$} for
the Week\_3 problem instance with a 40\% reduction.

\subsection{Reducing the Degree of Modification}
With regards the prior results, only the GA approach was able to
find solutions that always serviced all the scheduled tasks. The
proposed \emph{Partial-ACO} approaches both failed to scale up to
the larger problem instances. However, it was observed in the
experimental runs that diversity amongst the population became an
issue with ants traversing the same edges. Consequently, solutions
converged prematurely. A method that could be used to counter this
effect is to ensure that the degree to which an ant can modify its
$l_{best}$ solution is limited to a maximum percentage of the
solution. This should have the effect of preventing an ant from
copying the globally best tour. In fact, in the original
\emph{Partial-ACO} work, reducing the degree of modification of
solutions improved results and secondly, increased the speed of
the approach by reducing the probabilistic decision making of the
ants \cite{chitty:2017}.

\sisetup{ table-number-alignment=center,
separate-uncertainty=true, table-figures-integer = 2,
table-figures-decimal = 2}
\begin{table}
\footnotesize \centering \caption{The results from both
\emph{Partial-ACO} and \emph{Partial-ACO$^{PH}$} when applying a
maximum of 50\% modification of ant's locally best solution.
Results are shown in terms of the percentage of service time
demand achieved and the percentage reduction in traversal time of
the vehicle fleet over the original company
scheduling.\vspace{-0.25cm}}
\begin{tabular}{c
                S[separate-uncertainty,table-figures-uncertainty=1,table-figures-integer = 3, table-figures-decimal = 2, table-column-width=14mm]
                S[separate-uncertainty,table-figures-uncertainty=1,table-figures-integer = 2, table-figures-decimal = 2, table-column-width=13mm]
                S[separate-uncertainty,table-figures-uncertainty=1,table-figures-integer = 3, table-figures-decimal = 2, table-column-width=14mm]
                S[separate-uncertainty,table-figures-uncertainty=1,table-figures-integer = 2, table-figures-decimal = 2, table-column-width=12mm]}

 \toprule
{\multirow{4}{0.9cm}{Problem}}  & \multicolumn{2}{c}{Partial-ACO} & \multicolumn{2}{c}{Partial-ACO$^{PH}$}\\
\cmidrule{2-5} & {Job Time} & {Traversal} & {Job Time} & {Traversal}\\
& {Serviced} & {Reduction} & {Serviced} & {Reduction}\\
& {(\%)} & {(\%)} & {(\%)} & {(\%)}\\
\midrule
Week\_1 & 100.00 \pm 0.00 & 41.17 \pm 0.49 & 100.00 \pm 0.00 & 41.94 \pm 1.37 \\
Week\_2 & 100.00 \pm 0.00 & 42.39 \pm 1.11 & 100.00 \pm 0.00 & 43.48 \pm 0.44 \\
Week\_3 & 100.00 \pm 0.00 & 44.43 \pm 0.77 & 100.00 \pm 0.00 & 44.84 \pm 0.43 \\
Week\_4 & 100.00 \pm 0.00 & 38.55 \pm 1.58 & 100.00 \pm 0.00 & 40.78 \pm 0.41 \\
Fortnight\_1 & 100.00 \pm 0.00 & 23.89 \pm 6.54 & 100.00 \pm 0.00 & 34.30 \pm 2.39 \\
Fortnight\_2 & 100.00 \pm 0.00 & 24.41 \pm 5.13 & 100.00 \pm 0.00 & 34.08 \pm 1.12 \\
Fortnight\_3 & 100.00 \pm 0.00 & 35.55 \pm 9.35 & 100.00 \pm 0.00 & 41.32 \pm 1.89 \\
Fortnight\_4 & 100.00 \pm 0.00 & 31.07 \pm 11.24 & 100.00 \pm 0.00 & 37.51 \pm 1.57 \\
ThreeWeek\_1 & 100.00 \pm 0.00 & 28.16 \pm 5.05 & 100.00 \pm 0.00 & 21.78 \pm 2.46 \\
ThreeWeek\_2 & 100.00 \pm 0.00 & 23.54 \pm 4.48 & 100.00 \pm 0.00 & 21.00 \pm 2.40 \\
ThreeWeek\_3 & 100.00 \pm 0.00 & 27.12 \pm 4.17 & 100.00 \pm 0.00 & 25.94 \pm 1.87 \\
ThreeWeek\_4 & 100.00 \pm 0.00 & 22.52 \pm 4.05 & 100.00 \pm 0.00 & 22.29 \pm 1.53 \\
Month\_1 & 100.00 \pm 0.00 & 28.19 \pm 5.90 & 98.67 \pm 0.28 & -5.98 \pm 3.54 \\
\bottomrule
\end{tabular} \centering
\label{tab:ImprovedResults}
\end{table}

Table \ref{tab:ImprovedResults} demonstrates the results from both
the \emph{Partial-ACO} approaches when the degree of modification
of an ant's $l_{best}$ solution is restricted to a maximum of
50\%. However, to ensure there is scope to escape local optima
this restriction is removed with a probability of 0.001 during the
search process. From these results it can be observed that there
is a considerable improvement in the quality of the solutions
achieved by \emph{Partial-ACO} and it now scales to the larger
problem instances. Indeed, the technique now achieves results
considerably better than the GA approach for the week long and two
week long scenarios. However, for the larger problems the GA
meta-heuristic approach still achieves slightly better reductions
in fleet traversal times. With regards the pheromone matrix based
implementation, \emph{Partial-ACO$^{PH}$}, slightly better
solutions are achieved for the smaller problem instances compared
to the no pheromone matrix version. Reductions of up to 44\% are
achieved regards the fleet traversal times of the original
scheduling of the company that supplied the problem data.

\sisetup{ table-number-alignment=center,
separate-uncertainty=true, table-figures-integer = 2,
table-figures-decimal = 2}
\begin{table}
\footnotesize \centering \caption{The results from both
\emph{Partial-ACO} and \emph{Partial-ACO$^{PH}$} when applying a
maximum of 25\% modification of ant's locally best solution.
Results are shown in terms of the percentage of service time
demand achieved and the percentage reduction in traversal time of
the vehicle fleet over the original company
scheduling.\vspace{-0.25cm}}
\begin{tabular}{c
                S[separate-uncertainty,table-figures-uncertainty=1,table-figures-integer = 3, table-figures-decimal = 2, table-column-width=14mm]
                S[separate-uncertainty,table-figures-uncertainty=1,table-figures-integer = 2, table-figures-decimal = 2, table-column-width=13mm]
                S[separate-uncertainty,table-figures-uncertainty=1,table-figures-integer = 3, table-figures-decimal = 2, table-column-width=14mm]
                S[separate-uncertainty,table-figures-uncertainty=1,table-figures-integer = 2, table-figures-decimal = 2, table-column-width=12mm]}

 \toprule
{\multirow{4}{0.8cm}{Problem}}  & \multicolumn{2}{c}{Partial-ACO} & \multicolumn{2}{c}{Partial-ACO$^{PH}$}\\
\cmidrule{2-5} & {Job Time} & {Traversal} & {Job Time} & {Traversal}\\
& {Serviced} & {Reduction} & {Serviced} & {Reduction}\\
& {(\%)} & {(\%)} & {(\%)} & {(\%)}\\
\midrule
Week\_1 & 100.00 \pm 0.00 & 32.93 \pm 13.07 & 100.00 \pm 0.00 & 30.12 \pm 2.39 \\
Week\_2 & 100.00 \pm 0.00 & 33.91 \pm 5.24 & 100.00 \pm 0.00 & 30.84 \pm 4.91 \\
Week\_3 & 100.00 \pm 0.00 & 33.95 \pm 3.10 & 100.00 \pm 0.00 & 33.20 \pm 1.64 \\
Week\_4 & 100.00 \pm 0.00 & 35.06 \pm 3.28 & 100.00 \pm 0.00 & 30.16 \pm 3.49 \\
Fortnight\_1 & 100.00 \pm 0.00 & 28.08 \pm 7.53 & 100.00 \pm 0.00 & 31.38 \pm 2.37 \\
Fortnight\_2 & 100.00 \pm 0.00 & 30.66 \pm 6.14 & 100.00 \pm 0.00 & 32.15 \pm 2.67 \\
Fortnight\_3 & 100.00 \pm 0.00 & 27.27 \pm 9.15 & 100.00 \pm 0.00 & 35.77 \pm 2.33 \\
Fortnight\_4 & 100.00 \pm 0.00 & 33.50 \pm 7.27 & 100.00 \pm 0.00 & 30.24 \pm 2.42 \\
ThreeWeek\_1 & 100.00 \pm 0.00 & 31.32 \pm 6.41 & 100.00 \pm 0.00 & 23.45 \pm 3.21 \\
ThreeWeek\_2 & 100.00 \pm 0.00 & 30.23 \pm 5.22 & 100.00 \pm 0.00 & 19.84 \pm 2.03 \\
ThreeWeek\_3 & 100.00 \pm 0.00 & 33.27 \pm 6.85 & 100.00 \pm 0.00 & 26.18 \pm 2.89 \\
ThreeWeek\_4 & 100.00 \pm 0.00 & 31.66 \pm 6.74 & 100.00 \pm 0.00 & 22.78 \pm 2.42 \\
Month\_1 & 100.00 \pm 0.00 & 32.91 \pm 3.73 & 99.41 \pm 0.32 & 5.35 \pm 5.65 \\
\bottomrule
\end{tabular} \centering
\label{tab:Improved25}\vspace{-0.45cm}
\end{table}

This concept can be extended by further reducing the maximum
permissible degree of modification of an ant's $l_{best}$ solution
to just 25\%. Again though, to ensure there is scope to escape
local optima this restriction is removed with a probability of
0.001 during the search process. These results are shown in Table
\ref{tab:Improved25} whereby from the results regards
\emph{Partial-ACO} two effects can be observed. For the larger
problem instances the such as Month\_1 improved reductions in
traversal times are achieved of 30-33\% which when now compared
the to the GA approach are several percent better. However, with
the smaller one week problems the results are poorer. This is
because with the larger problems there is still quite a few
vehicles and jobs that can be reorganised even with only a 25\%
maximum modification restriction. But for the smaller problems
there is not much capacity for change hence the worsening results.
Therefore, it can be postulated that restricting the maximum
modification is beneficial to the performance of
\emph{Partial-ACO} the size of the problem needs to be taken into
account such there is sufficient scope for sufficient solution
modification. With regards \emph{Partial-ACO$^{PH}$} only minor
improvements are noted for the larger problems and a similar
degradation in performance for the smaller problems. It should be
noted that all meta-heuristics did not consistently find an
optimal solution. However, all results were generated without any
local search operator such as 2-opt which would likely
significantly improve the results.

\subsection{Execution Timings}
One final aspect of analysis left to consider is the execution
timings of the various approaches evaluated and these are shown in
Table \ref{tab:ExecutionTimings}. The GA approach is clearly the
fastest approach over three times faster than the ACO MMAS
approach and nearly four times faster than the \emph{Partial-ACO}
approaches. This is because the GA approach does not need to
construct solutions by probabilistically investigating each every
option at each step. Furthermore, \emph{Partial-ACO} is required
to reconstruct pheromone on the unvisited edges from the
population. \emph{Partial-ACO$^{PH}$} is equally slow as this
approach needs to update the pheromone matrix each time all ants
have constructed a new solution. However, Table
\ref{tab:ExecutionTimings5025} shows the execution timings for the
\emph{Partial-ACO} approaches when restricting the degree of
modification. For \emph{Partial-ACO} execution timings have
reduced considerably, although less than expected, through being
able to make fewer pheromone reconstructions as less decisions are
required to construct a solution. Execution timings are now on a
par with a GA approach. With regards \emph{Partial-ACO$^{PH}$}
reductions in execution times are also observed but to a lower
degree since the approach still has to update the pheromone
matrix.

\sisetup{ table-number-alignment=center,
separate-uncertainty=true, table-figures-integer = 2,
table-figures-decimal = 2}
\begin{table}
\footnotesize \centering \caption{The execution timings in minutes
across all of the meta-heuristics tested.\vspace{-0.25cm}}
\begin{tabular}{c
                S[separate-uncertainty,table-figures-uncertainty=1,table-figures-integer = 1, table-figures-decimal = 2, table-column-width=12mm]
                S[separate-uncertainty,table-figures-uncertainty=1,table-figures-integer = 2, table-figures-decimal = 2, table-column-width=12mm]
                S[separate-uncertainty,table-figures-uncertainty=1,table-figures-integer = 2, table-figures-decimal = 2, table-column-width=14mm]
                S[separate-uncertainty,table-figures-uncertainty=1,table-figures-integer = 2, table-figures-decimal = 2, table-column-width=14mm]}

 \toprule
{Problem} & {GA} & {MMAS} & {Partial-ACO} & {Partial-ACO$^{PH}$}\\
\midrule
Week\_1 & 1.32 \pm 0.04 & 2.23 \pm 0.10 & 4.69 \pm 0.49 & 3.61 \pm 0.02 \\
Week\_2 & 1.35 \pm 0.04 & 2.33 \pm 0.10 & 4.76 \pm 0.45 & 3.69 \pm 0.03 \\
Week\_3 & 1.35 \pm 0.04 & 2.49 \pm 0.10 & 4.90 \pm 0.50 & 3.82 \pm 0.03 \\
Week\_4 & 1.09 \pm 0.03 & 1.64 \pm 0.05 & 3.86 \pm 0.51 & 2.85 \pm 0.02 \\
Fortnight\_1 & 2.52 \pm 0.03 & 6.56 \pm 0.13 & 9.87 \pm 0.43 & 8.92 \pm 0.04 \\
Fortnight\_2 & 2.20 \pm 0.02 & 5.52 \pm 0.07 & 8.81 \pm 0.50 & 7.72 \pm 0.06 \\
Fortnight\_3 & 2.59 \pm 0.02 & 6.84 \pm 0.10 & 10.18 \pm 0.38 & 9.26 \pm 0.07 \\
Fortnight\_4 & 2.26 \pm 0.02 & 5.76 \pm 0.11 & 9.09 \pm 0.48 & 7.96 \pm 0.05 \\
ThreeWeek\_1 & 4.30 \pm 0.03 & 13.09 \pm 0.11 & 16.42 \pm 0.71 & 16.34 \pm 0.03 \\
ThreeWeek\_2 & 3.95 \pm 0.02 & 11.79 \pm 0.08 & 14.81 \pm 0.68 & 14.51 \pm 0.06 \\
ThreeWeek\_3 & 3.97 \pm 0.02 & 11.57 \pm 0.15 & 14.54 \pm 0.78 & 14.65 \pm 0.05 \\
ThreeWeek\_4 & 4.06 \pm 0.03 & 12.09 \pm 0.15 & 14.95 \pm 0.49 & 14.85 \pm 0.06 \\
Month\_1 & 5.96 \pm 0.04 & 19.75 \pm 0.13 & 23.59 \pm 0.83 & 23.18 \pm 0.05 \\
\bottomrule
\end{tabular} \centering
\label{tab:ExecutionTimings}\vspace{-0.35cm}
\end{table}

\vspace{-0.1cm}
\section{Conclusions}
The work presented in this paper has investigated the use of
meta-heuristic approaches for tackling larger scale fleet
optimisation problems. Specifically, the use of the novel
\emph{Partial-ACO} approach has been compared to the main
competing meta-heuristic approaches, GA and ACO.
\emph{Partial-ACO} is similar to ACO but involves ants maintaining
a memory of their best found solution and only partially modifying
this solution at each iteration. The premise behind this technique
was that ACO will struggle to scale well to larger problems
because as the size of the problem increases the potential for an
ant to probabilistically make an error also increases. By reducing
the number of decisions an ant makes when creating a new solution,
the probability of an ant making a poor choice is similarly
reduced enabling ACO to scale to larger problems better.

\sisetup{ table-number-alignment=center,
separate-uncertainty=true, table-figures-integer = 2,
table-figures-decimal = 2}
\begin{table}
\footnotesize \centering \caption{The execution timings in minutes
for the \emph{Partial-ACO} approaches and restrictions on solution
modification. \vspace{-0.5cm}}
\begin{tabular}{c
                S[separate-uncertainty,table-figures-uncertainty=1,table-figures-integer = 2, table-figures-decimal = 2, table-column-width=12mm]
                S[separate-uncertainty,table-figures-uncertainty=1,table-figures-integer = 2, table-figures-decimal = 2, table-column-width=15mm]
                S[separate-uncertainty,table-figures-uncertainty=1,table-figures-integer = 2, table-figures-decimal = 2, table-column-width=12mm]
                S[separate-uncertainty,table-figures-uncertainty=1,table-figures-integer = 2, table-figures-decimal = 2, table-column-width=14mm]}

 \toprule
{\multirow{2}{0.9cm}{Problem}}  & \multicolumn{2}{c}{Max. Modification 50\%} & \multicolumn{2}{c}{Max. Modification 25\%}\\
\cmidrule{2-5} & {Partial-ACO} & {Partial-ACO$^{PH}$} & {Partial-ACO} & {Partial-ACO$^{PH}$}\\
\midrule
Week\_1 & 2.96 \pm 0.64 & 3.33 \pm 0.02 & 2.03 \pm 0.32 & 3.25 \pm 0.05 \\
Week\_2 & 2.96 \pm 0.73 & 3.38 \pm 0.02 & 2.02 \pm 0.17 & 3.25 \pm 0.05 \\
Week\_3 & 3.07 \pm 0.59 & 3.52 \pm 0.03 & 2.05 \pm 0.05 & 3.38 \pm 0.05 \\
Week\_4 & 2.91 \pm 0.84 & 2.66 \pm 0.05 & 1.67 \pm 0.04 & 2.53 \pm 0.05 \\
Fortnight\_1 & 5.56 \pm 0.36 & 7.81 \pm 0.04 & 3.36 \pm 0.40 & 7.21 \pm 0.05 \\
Fortnight\_2 & 4.88 \pm 0.42 & 6.76 \pm 0.03 & 3.08 \pm 0.12 & 6.31 \pm 0.06 \\
Fortnight\_3 & 6.03 \pm 0.29 & 8.04 \pm 0.05 & 3.50 \pm 0.08 & 7.45 \pm 0.07 \\
Fortnight\_4 & 5.15 \pm 0.38 & 6.97 \pm 0.03 & 3.20 \pm 0.06 & 6.50 \pm 0.06 \\
ThreeWeek\_1 & 8.20 \pm 0.49 & 13.80 \pm 0.06 & 4.93 \pm 0.35 & 12.60 \pm 0.07 \\
ThreeWeek\_2 & 7.38 \pm 0.48 & 12.41 \pm 0.06 & 4.62 \pm 0.38 & 11.35 \pm 0.07 \\
ThreeWeek\_3 & 7.45 \pm 0.50 & 12.79 \pm 0.09 & 4.58 \pm 0.54 & 11.67 \pm 0.05 \\
ThreeWeek\_4 & 7.69 \pm 0.44 & 12.75 \pm 0.11 & 4.60 \pm 0.34 & 11.59 \pm 0.06 \\
Month\_1 & 10.64 \pm 0.52 & 19.40 \pm 0.08 & 6.18 \pm 0.24 & 17.64 \pm 0.12 \\
\bottomrule
\end{tabular} \centering
\label{tab:ExecutionTimings5025}\vspace{-0.45cm}
\end{table}

The \emph{Partial-ACO} approach with and without a pheromone
matrix was tested on a number of real-world problems of increasing
complexity and size and compared with the more familiar GA and ACO
meta-heuristics. Overall, \emph{Partial-ACO} proved competitive
with a GA approach and significantly better than the ACO approach
both with and without a pheromone matrix. Indeed, these results
reinforced the hypothesis that the more decisions an ant has to
make when constructing a solution the greater the probability of
making an error occurs reducing the capability for ACO to scale to
larger problems. However, it was evidenced that \emph{Partial-ACO}
can bypass this potential problem. By ants being able to make
fewer decisions as a result of only partially modifying an
existing good solution, the probability of making a poor decision
is reduced leading to higher quality solutions. Moreover, through
a lower degree of decision making the computational speed of
\emph{Partial-ACO} is much improved. Overall, \emph{Partial-ACO}
was able to improve upon the schedules provided by the Birmingham
based company which provided the dataset by up to 44\% for smaller
problems and 28\% for the largest problem containing 32 vehicles
and 298 jobs.

It can be stated that fleet optimisation using \emph{Partial-ACO}
is capable of providing significant cost savings to commercial
companies. Moreover, reducing vehicle traversal to this degree
provides significant assistance to reducing vehicular pollution in
cities at little to no cost. Future work will consist of
developing better strategies with regards the solution
preservation aspect of \emph{Partial-ACO} and a complete
investigation of parameter settings with a view to scaling
\emph{Partial-ACO} to even larger fleet optimisation problems.

\vspace{-0.1cm}
\section{Acknowledgement} This work
was carried out under the System Analytics for Innovation project,
which is part-funded by the European Regional Development Fund
(ERDF).

\bibliographystyle{ACM-Reference-Format}
\bibliography{PartialACOforFleetOptimisation}


\begin{thebibliography}{46}


\ifx \showCODEN    \undefined \def \showCODEN     #1{\unskip}     \fi
\ifx \showDOI      \undefined \def \showDOI       #1{#1}\fi
\ifx \showISBNx    \undefined \def \showISBNx     #1{\unskip}     \fi
\ifx \showISBNxiii \undefined \def \showISBNxiii  #1{\unskip}     \fi
\ifx \showISSN     \undefined \def \showISSN      #1{\unskip}     \fi
\ifx \showLCCN     \undefined \def \showLCCN      #1{\unskip}     \fi
\ifx \shownote     \undefined \def \shownote      #1{#1}          \fi
\ifx \showarticletitle \undefined \def \showarticletitle #1{#1}   \fi
\ifx \showURL      \undefined \def \showURL       {\relax}        \fi
\providecommand\bibfield[2]{#2}
\providecommand\bibinfo[2]{#2}
\providecommand\natexlab[1]{#1}
\providecommand\showeprint[2][]{arXiv:#2}

\bibitem[\protect\citeauthoryear{Alba and Dorronsoro}{Alba and
  Dorronsoro}{2006}]%
        {alba:2006}
\bibfield{author}{\bibinfo{person}{Enrique Alba} {and}
  \bibinfo{person}{Bernab{\'e} Dorronsoro}.} \bibinfo{year}{2006}\natexlab{}.
\newblock \showarticletitle{Computing nine new best-so-far solutions for
  capacitated {VRP} with a cellular genetic algorithm}.
\newblock \bibinfo{journal}{{\it Inform. Process. Lett.}} \bibinfo{volume}{98},
  \bibinfo{number}{6} (\bibinfo{year}{2006}), \bibinfo{pages}{225--230}.
\newblock


\bibitem[\protect\citeauthoryear{Benavent and Mart{\'\i}nez}{Benavent and
  Mart{\'\i}nez}{2013}]%
        {benavent:2013}
\bibfield{author}{\bibinfo{person}{Enrique Benavent} {and}
  \bibinfo{person}{Antonio Mart{\'\i}nez}.} \bibinfo{year}{2013}\natexlab{}.
\newblock \showarticletitle{Multi-depot multiple {TSP}: a polyhedral study and
  computational results}.
\newblock \bibinfo{journal}{{\em Annals of Operations Research\/}}
  \bibinfo{volume}{207}, \bibinfo{number}{1} (\bibinfo{year}{2013}),
  \bibinfo{pages}{7--25}.
\newblock


\bibitem[\protect\citeauthoryear{Braekers, Caris, and Janssens}{Braekers
  et~al\mbox{.}}{2014}]%
        {braekers:2014}
\bibfield{author}{\bibinfo{person}{Kris Braekers}, \bibinfo{person}{An Caris},
  {and} \bibinfo{person}{Gerrit~K Janssens}.} \bibinfo{year}{2014}\natexlab{}.
\newblock \showarticletitle{Exact and meta-heuristic approach for a general
  heterogeneous dial-a-ride problem with multiple depots}.
\newblock \bibinfo{journal}{{\em Transportation Research Part B:
  Methodological\/}}  \bibinfo{volume}{67} (\bibinfo{year}{2014}),
  \bibinfo{pages}{166--186}.
\newblock


\bibitem[\protect\citeauthoryear{Calder{\'o}n-Garcidue{\~n}as, Leray,
  Heydarpour, Torres-Jard{\'o}n, and Reis}{Calder{\'o}n-Garcidue{\~n}as
  et~al\mbox{.}}{2016}]%
        {calderon:2016}
\bibfield{author}{\bibinfo{person}{L Calder{\'o}n-Garcidue{\~n}as},
  \bibinfo{person}{E Leray}, \bibinfo{person}{P Heydarpour}, \bibinfo{person}{R
  Torres-Jard{\'o}n}, {and} \bibinfo{person}{J Reis}.}
  \bibinfo{year}{2016}\natexlab{}.
\newblock \showarticletitle{Air pollution, a rising environmental risk factor
  for cognition, neuroinflammation and neurodegeneration: the clinical impact
  on children and beyond}.
\newblock \bibinfo{journal}{{\em Revue neurologique\/}} \bibinfo{volume}{172},
  \bibinfo{number}{1} (\bibinfo{year}{2016}), \bibinfo{pages}{69--80}.
\newblock


\bibitem[\protect\citeauthoryear{Calvete, Gal{\'e}, and Oliveros}{Calvete
  et~al\mbox{.}}{2011}]%
        {calvete:2011}
\bibfield{author}{\bibinfo{person}{Herminia~I Calvete}, \bibinfo{person}{Carmen
  Gal{\'e}}, {and} \bibinfo{person}{Mar{\'\i}a-Jos{\'e} Oliveros}.}
  \bibinfo{year}{2011}\natexlab{}.
\newblock \showarticletitle{Evolutive and {ACO} Strategies for Solving the
  Multi-depot Vehicle Routing Problem.}. In \bibinfo{booktitle}{{\em IJCCI
  (ECTA-FCTA)}}. \bibinfo{pages}{73--79}.
\newblock


\bibitem[\protect\citeauthoryear{Chao, Golden, and Wasil}{Chao
  et~al\mbox{.}}{1993}]%
        {chao:1993}
\bibfield{author}{\bibinfo{person}{I-Ming Chao}, \bibinfo{person}{Bruce~L
  Golden}, {and} \bibinfo{person}{Edward Wasil}.}
  \bibinfo{year}{1993}\natexlab{}.
\newblock \showarticletitle{A new heuristic for the multi-depot vehicle routing
  problem that improves upon best-known solutions}.
\newblock \bibinfo{journal}{{\em American Journal of Mathematical and
  Management Sciences\/}} \bibinfo{volume}{13}, \bibinfo{number}{3-4}
  (\bibinfo{year}{1993}), \bibinfo{pages}{371--406}.
\newblock


\bibitem[\protect\citeauthoryear{Chitty}{Chitty}{2017}]%
        {chitty:2017}
\bibfield{author}{\bibinfo{person}{Darren~M Chitty}.}
  \bibinfo{year}{2017}\natexlab{}.
\newblock \showarticletitle{Applying {ACO} to Large Scale {TSP} Instances}. In
  \bibinfo{booktitle}{{\em UK Workshop on Computational Intelligence}}.
  Springer, \bibinfo{pages}{104--118}.
\newblock


\bibitem[\protect\citeauthoryear{Clarke and Wright}{Clarke and Wright}{1964}]%
        {clarke:1964}
\bibfield{author}{\bibinfo{person}{Geoff Clarke} {and} \bibinfo{person}{John~W
  Wright}.} \bibinfo{year}{1964}\natexlab{}.
\newblock \showarticletitle{Scheduling of vehicles from a central depot to a
  number of delivery points}.
\newblock \bibinfo{journal}{{\em Operations research\/}} \bibinfo{volume}{12},
  \bibinfo{number}{4} (\bibinfo{year}{1964}), \bibinfo{pages}{568--581}.
\newblock


\bibitem[\protect\citeauthoryear{Dantzig and Ramser}{Dantzig and
  Ramser}{1959}]%
        {dantzig:1959}
\bibfield{author}{\bibinfo{person}{George~B Dantzig} {and}
  \bibinfo{person}{John~H Ramser}.} \bibinfo{year}{1959}\natexlab{}.
\newblock \showarticletitle{The truck dispatching problem}.
\newblock \bibinfo{journal}{{\em Management science\/}} \bibinfo{volume}{6},
  \bibinfo{number}{1} (\bibinfo{year}{1959}), \bibinfo{pages}{80--91}.
\newblock


\bibitem[\protect\citeauthoryear{Dondo, Mendez, and Cerd{\'a}}{Dondo
  et~al\mbox{.}}{2003}]%
        {dondo:2003}
\bibfield{author}{\bibinfo{person}{Rodolfo Dondo},
  \bibinfo{person}{Carlos~Alberto Mendez}, {and} \bibinfo{person}{Jaime
  Cerd{\'a}}.} \bibinfo{year}{2003}\natexlab{}.
\newblock \showarticletitle{An optimal approach to the multiple-depot
  heterogeneous vehicle routing problem with time window and capacity
  constraints}.
\newblock \bibinfo{journal}{{\em Latin American applied research\/}}
  \bibinfo{volume}{33}, \bibinfo{number}{2} (\bibinfo{year}{2003}),
  \bibinfo{pages}{129--134}.
\newblock


\bibitem[\protect\citeauthoryear{Dondo, M{\'e}ndez, and Cerd{\'a}}{Dondo
  et~al\mbox{.}}{2008}]%
        {dondo:2008}
\bibfield{author}{\bibinfo{person}{Rodolfo Dondo}, \bibinfo{person}{Carlos~A
  M{\'e}ndez}, {and} \bibinfo{person}{Jaime Cerd{\'a}}.}
  \bibinfo{year}{2008}\natexlab{}.
\newblock \showarticletitle{Optimal management of logistic activities in
  multi-site environments}.
\newblock \bibinfo{journal}{{\em Computers \& Chemical Engineering\/}}
  \bibinfo{volume}{32}, \bibinfo{number}{11} (\bibinfo{year}{2008}),
  \bibinfo{pages}{2547--2569}.
\newblock


\bibitem[\protect\citeauthoryear{Dondo and Cerd{\'a}}{Dondo and
  Cerd{\'a}}{2009}]%
        {dondo:2009}
\bibfield{author}{\bibinfo{person}{Rodolfo~G Dondo} {and}
  \bibinfo{person}{Jaime Cerd{\'a}}.} \bibinfo{year}{2009}\natexlab{}.
\newblock \showarticletitle{A hybrid local improvement algorithm for
  large-scale multi-depot vehicle routing problems with time windows}.
\newblock \bibinfo{journal}{{\em Computers \& Chemical Engineering\/}}
  \bibinfo{volume}{33}, \bibinfo{number}{2} (\bibinfo{year}{2009}),
  \bibinfo{pages}{513--530}.
\newblock


\bibitem[\protect\citeauthoryear{Dorigo and Gambardella}{Dorigo and
  Gambardella}{1997}]%
        {dorigo:1997}
\bibfield{author}{\bibinfo{person}{Marco Dorigo} {and}
  \bibinfo{person}{Luca~Maria Gambardella}.} \bibinfo{year}{1997}\natexlab{}.
\newblock \showarticletitle{Ant colony system: a cooperative learning approach
  to the traveling salesman problem}.
\newblock \bibinfo{journal}{{\em IEEE Transactions on evolutionary
  computation\/}} \bibinfo{volume}{1}, \bibinfo{number}{1}
  (\bibinfo{year}{1997}), \bibinfo{pages}{53--66}.
\newblock


\bibitem[\protect\citeauthoryear{Eberhart and Kennedy}{Eberhart and
  Kennedy}{1995}]%
        {eberhart:1995}
\bibfield{author}{\bibinfo{person}{Russell Eberhart} {and}
  \bibinfo{person}{James Kennedy}.} \bibinfo{year}{1995}\natexlab{}.
\newblock \showarticletitle{A new optimizer using particle swarm theory}. In
  \bibinfo{booktitle}{{\em Micro Machine and Human Science, 1995. MHS'95.,
  Proceedings of the Sixth International Symposium on}}. IEEE,
  \bibinfo{pages}{39--43}.
\newblock


\bibitem[\protect\citeauthoryear{Filipec, Skrlec, and Krajcar}{Filipec
  et~al\mbox{.}}{1997}]%
        {filipec:1997}
\bibfield{author}{\bibinfo{person}{Minea Filipec}, \bibinfo{person}{Davor
  Skrlec}, {and} \bibinfo{person}{Slavko Krajcar}.}
  \bibinfo{year}{1997}\natexlab{}.
\newblock \showarticletitle{Darwin meets computers: New approach to multiple
  depot capacitated vehicle routing problem}. In \bibinfo{booktitle}{{\em
  Systems, Man, and Cybernetics, 1997. Computational Cybernetics and
  Simulation., 1997 IEEE International Conference on}},
  Vol.~\bibinfo{volume}{1}. IEEE, \bibinfo{pages}{421--426}.
\newblock


\bibitem[\protect\citeauthoryear{Geetha, Poonthalir, and Vanathi}{Geetha
  et~al\mbox{.}}{2013}]%
        {geetha:2013}
\bibfield{author}{\bibinfo{person}{S Geetha}, \bibinfo{person}{G Poonthalir},
  {and} \bibinfo{person}{PT Vanathi}.} \bibinfo{year}{2013}\natexlab{}.
\newblock \showarticletitle{Nested particle swarm optimisation for multi-depot
  vehicle routing problem}.
\newblock \bibinfo{journal}{{\em International Journal of Operational
  Research\/}} \bibinfo{volume}{16}, \bibinfo{number}{3}
  (\bibinfo{year}{2013}), \bibinfo{pages}{329--348}.
\newblock


\bibitem[\protect\citeauthoryear{Geetha, Vanathi, and Poonthalir}{Geetha
  et~al\mbox{.}}{2012}]%
        {geetha:2012}
\bibfield{author}{\bibinfo{person}{Shanmugam Geetha}, \bibinfo{person}{PT
  Vanathi}, {and} \bibinfo{person}{Ganesan Poonthalir}.}
  \bibinfo{year}{2012}\natexlab{}.
\newblock \showarticletitle{Metaheuristic approach for the multi-depot vehicle
  routing problem}.
\newblock \bibinfo{journal}{{\em Applied Artificial Intelligence\/}}
  \bibinfo{volume}{26}, \bibinfo{number}{9} (\bibinfo{year}{2012}),
  \bibinfo{pages}{878--901}.
\newblock


\bibitem[\protect\citeauthoryear{Gendreau, Hertz, and Laporte}{Gendreau
  et~al\mbox{.}}{1994}]%
        {gendreau:1994}
\bibfield{author}{\bibinfo{person}{Michel Gendreau}, \bibinfo{person}{Alain
  Hertz}, {and} \bibinfo{person}{Gilbert Laporte}.}
  \bibinfo{year}{1994}\natexlab{}.
\newblock \showarticletitle{A tabu search heuristic for the vehicle routing
  problem}.
\newblock \bibinfo{journal}{{\em Management science\/}} \bibinfo{volume}{40},
  \bibinfo{number}{10} (\bibinfo{year}{1994}), \bibinfo{pages}{1276--1290}.
\newblock


\bibitem[\protect\citeauthoryear{Gilbert}{Gilbert}{1992}]%
        {laporte:1992}
\bibfield{author}{\bibinfo{person}{Laporte Gilbert}.}
  \bibinfo{year}{1992}\natexlab{}.
\newblock \showarticletitle{The vehicle routing problem: An overview of exact
  and approximate algorithms}.
\newblock \bibinfo{journal}{{\em European journal of operational research\/}}
  \bibinfo{volume}{59}, \bibinfo{number}{3} (\bibinfo{year}{1992}),
  \bibinfo{pages}{345--358}.
\newblock


\bibitem[\protect\citeauthoryear{Gilbert, Desrochers, and Norbert}{Gilbert
  et~al\mbox{.}}{1984}]%
        {laporte:1984}
\bibfield{author}{\bibinfo{person}{Laporte Gilbert}, \bibinfo{person}{Martin
  Desrochers}, {and} \bibinfo{person}{Yves Norbert}.}
  \bibinfo{year}{1984}\natexlab{}.
\newblock \showarticletitle{Two exact algorithms for the distance-constrained
  vehicle routing problem}.
\newblock \bibinfo{journal}{{\em Networks\/}} \bibinfo{volume}{14},
  \bibinfo{number}{1} (\bibinfo{year}{1984}), \bibinfo{pages}{161--172}.
\newblock


\bibitem[\protect\citeauthoryear{Gillett and Johnson}{Gillett and
  Johnson}{1976}]%
        {gillett:1976}
\bibfield{author}{\bibinfo{person}{Billy~E Gillett} {and}
  \bibinfo{person}{Jerry~G Johnson}.} \bibinfo{year}{1976}\natexlab{}.
\newblock \showarticletitle{Multi-terminal vehicle-dispatch algorithm}.
\newblock \bibinfo{journal}{{\em Omega\/}} \bibinfo{volume}{4},
  \bibinfo{number}{6} (\bibinfo{year}{1976}), \bibinfo{pages}{711--718}.
\newblock


\bibitem[\protect\citeauthoryear{Golden, Magnanti, and Nguyen}{Golden
  et~al\mbox{.}}{1977}]%
        {golden:1977}
\bibfield{author}{\bibinfo{person}{Bruce~L Golden}, \bibinfo{person}{Thomas~L
  Magnanti}, {and} \bibinfo{person}{Hien~Q Nguyen}.}
  \bibinfo{year}{1977}\natexlab{}.
\newblock \showarticletitle{Implementing vehicle routing algorithms}.
\newblock \bibinfo{journal}{{\em Networks\/}} \bibinfo{volume}{7},
  \bibinfo{number}{2} (\bibinfo{year}{1977}), \bibinfo{pages}{113--148}.
\newblock


\bibitem[\protect\citeauthoryear{Guntsch and Middendorf}{Guntsch and
  Middendorf}{2002}]%
        {guntsch:2002}
\bibfield{author}{\bibinfo{person}{Michael Guntsch} {and}
  \bibinfo{person}{Martin Middendorf}.} \bibinfo{year}{2002}\natexlab{}.
\newblock \showarticletitle{A population based approach for {ACO}}. In
  \bibinfo{booktitle}{{\em Workshops on Applications of Evolutionary
  Computation}}. Springer, \bibinfo{pages}{72--81}.
\newblock


\bibitem[\protect\citeauthoryear{Ho, Ho, Ji, and Lau}{Ho et~al\mbox{.}}{2008}]%
        {ho:2008}
\bibfield{author}{\bibinfo{person}{William Ho}, \bibinfo{person}{George~TS Ho},
  \bibinfo{person}{Ping Ji}, {and} \bibinfo{person}{Henry~CW Lau}.}
  \bibinfo{year}{2008}\natexlab{}.
\newblock \showarticletitle{A hybrid genetic algorithm for the multi-depot
  vehicle routing problem}.
\newblock \bibinfo{journal}{{\em Engineering Applications of Artificial
  Intelligence\/}} \bibinfo{volume}{21}, \bibinfo{number}{4}
  (\bibinfo{year}{2008}), \bibinfo{pages}{548--557}.
\newblock


\bibitem[\protect\citeauthoryear{Holland}{Holland}{1975}]%
        {Holland:1975}
\bibfield{author}{\bibinfo{person}{John~H Holland}.}
  \bibinfo{year}{1975}\natexlab{}.
\newblock \bibinfo{booktitle}{{\em Adaptation in natural and artificial
  systems: an introductory analysis with applications to biology, control, and
  artificial intelligence.}}
\newblock \bibinfo{publisher}{U Michigan Press}.
\newblock


\bibitem[\protect\citeauthoryear{Karakati{\v{c}} and
  Podgorelec}{Karakati{\v{c}} and Podgorelec}{2015}]%
        {karakativc:2015}
\bibfield{author}{\bibinfo{person}{Sa{\v{s}}o Karakati{\v{c}}} {and}
  \bibinfo{person}{Vili Podgorelec}.} \bibinfo{year}{2015}\natexlab{}.
\newblock \showarticletitle{A survey of genetic algorithms for solving multi
  depot vehicle routing problem}.
\newblock \bibinfo{journal}{{\em Applied Soft Computing\/}}
  \bibinfo{volume}{27} (\bibinfo{year}{2015}), \bibinfo{pages}{519--532}.
\newblock


\bibitem[\protect\citeauthoryear{Laporte, Nobert, and Taillefer}{Laporte
  et~al\mbox{.}}{1988}]%
        {laporte:1988}
\bibfield{author}{\bibinfo{person}{Gilbert Laporte}, \bibinfo{person}{Yves
  Nobert}, {and} \bibinfo{person}{Serge Taillefer}.}
  \bibinfo{year}{1988}\natexlab{}.
\newblock \showarticletitle{Solving a family of multi-depot vehicle routing and
  location-routing problems}.
\newblock \bibinfo{journal}{{\em Transportation science\/}}
  \bibinfo{volume}{22}, \bibinfo{number}{3} (\bibinfo{year}{1988}),
  \bibinfo{pages}{161--172}.
\newblock


\bibitem[\protect\citeauthoryear{Raft}{Raft}{1982}]%
        {raft:1982}
\bibfield{author}{\bibinfo{person}{Ole~M Raft}.}
  \bibinfo{year}{1982}\natexlab{}.
\newblock \showarticletitle{A modular algorithm for an extended vehicle
  scheduling problem}.
\newblock \bibinfo{journal}{{\em European Journal of Operational Research\/}}
  \bibinfo{volume}{11}, \bibinfo{number}{1} (\bibinfo{year}{1982}),
  \bibinfo{pages}{67--76}.
\newblock


\bibitem[\protect\citeauthoryear{Renaud, Laporte, and Boctor}{Renaud
  et~al\mbox{.}}{1996}]%
        {renaud:1996}
\bibfield{author}{\bibinfo{person}{Jacques Renaud}, \bibinfo{person}{Gilbert
  Laporte}, {and} \bibinfo{person}{Fayez~F Boctor}.}
  \bibinfo{year}{1996}\natexlab{}.
\newblock \showarticletitle{A tabu search heuristic for the multi-depot vehicle
  routing problem}.
\newblock \bibinfo{journal}{{\em Computers \& Operations Research\/}}
  \bibinfo{volume}{23}, \bibinfo{number}{3} (\bibinfo{year}{1996}),
  \bibinfo{pages}{229--235}.
\newblock


\bibitem[\protect\citeauthoryear{Requia, Adams, Arain, Papatheodorou,
  Koutrakis, and Mahmoud}{Requia et~al\mbox{.}}{2018}]%
        {requia:2018}
\bibfield{author}{\bibinfo{person}{Weeberb~J Requia},
  \bibinfo{person}{Matthew~D Adams}, \bibinfo{person}{Altaf Arain},
  \bibinfo{person}{Stefania Papatheodorou}, \bibinfo{person}{Petros Koutrakis},
  {and} \bibinfo{person}{Moataz Mahmoud}.} \bibinfo{year}{2018}\natexlab{}.
\newblock \showarticletitle{Global Association of air Pollution and
  Cardiorespiratory Diseases: a systematic review, meta-analysis, and
  investigation of modifier variables}.
\newblock \bibinfo{journal}{{\em American journal of public health\/}}
  \bibinfo{volume}{108}, \bibinfo{number}{S2} (\bibinfo{year}{2018}),
  \bibinfo{pages}{S123--S130}.
\newblock


\bibitem[\protect\citeauthoryear{Salhi and Nagy}{Salhi and Nagy}{1999}]%
        {salhi:1999}
\bibfield{author}{\bibinfo{person}{Sa{\"\i}d Salhi} {and}
  \bibinfo{person}{G{\'a}bor Nagy}.} \bibinfo{year}{1999}\natexlab{}.
\newblock \showarticletitle{A cluster insertion heuristic for single and
  multiple depot vehicle routing problems with backhauling}.
\newblock \bibinfo{journal}{{\em Journal of the operational Research
  Society\/}} \bibinfo{volume}{50}, \bibinfo{number}{10}
  (\bibinfo{year}{1999}), \bibinfo{pages}{1034--1042}.
\newblock


\bibitem[\protect\citeauthoryear{Salhi and Sari}{Salhi and Sari}{1997}]%
        {salhi:1997}
\bibfield{author}{\bibinfo{person}{Said Salhi} {and} \bibinfo{person}{M Sari}.}
  \bibinfo{year}{1997}\natexlab{}.
\newblock \showarticletitle{A multi-level composite heuristic for the
  multi-depot vehicle fleet mix problem}.
\newblock \bibinfo{journal}{{\em European Journal of Operational Research\/}}
  \bibinfo{volume}{103}, \bibinfo{number}{1} (\bibinfo{year}{1997}),
  \bibinfo{pages}{95--112}.
\newblock


\bibitem[\protect\citeauthoryear{Salhi, Thangiah, and Rahman}{Salhi
  et~al\mbox{.}}{1998}]%
        {salhi:1998}
\bibfield{author}{\bibinfo{person}{Said Salhi},
  \bibinfo{person}{Sam~Rabindranath Thangiah}, {and} \bibinfo{person}{Fuad
  Rahman}.} \bibinfo{year}{1998}\natexlab{}.
\newblock \showarticletitle{A genetic clustering method for the multi-depot
  vehicle routing problem}. In \bibinfo{booktitle}{{\em Artificial Neural Nets
  and Genetic Algorithms}}. Springer, \bibinfo{pages}{234--237}.
\newblock


\bibitem[\protect\citeauthoryear{Skok, Skrlec, and Krajcar}{Skok
  et~al\mbox{.}}{2000}]%
        {skok:2000}
\bibfield{author}{\bibinfo{person}{Minea Skok}, \bibinfo{person}{Davor Skrlec},
  {and} \bibinfo{person}{Slavko Krajcar}.} \bibinfo{year}{2000}\natexlab{}.
\newblock \showarticletitle{The non-fixed destination multiple depot
  capacitated vehicle routing problem and genetic algorithms}. In
  \bibinfo{booktitle}{{\em Information Technology Interfaces, 2000. ITI 2000.
  Proceedings of the 22nd International Conference on}}. IEEE,
  \bibinfo{pages}{403--408}.
\newblock


\bibitem[\protect\citeauthoryear{Skok, Skrlec, and Krajcar}{Skok
  et~al\mbox{.}}{2001}]%
        {skok:2001}
\bibfield{author}{\bibinfo{person}{Minea Skok}, \bibinfo{person}{Davor Skrlec},
  {and} \bibinfo{person}{Slavko Krajcar}.} \bibinfo{year}{2001}\natexlab{}.
\newblock \showarticletitle{The Genetic Algorithm Scheduling of Vehicles from
  Multiple Depots to a Number of Delivery Points}.
\newblock \bibinfo{journal}{{\em Arficial Intelligence\/}}
  \bibinfo{volume}{349} (\bibinfo{year}{2001}).
\newblock


\bibitem[\protect\citeauthoryear{Stodola}{Stodola}{2018}]%
        {stodola:2018}
\bibfield{author}{\bibinfo{person}{Petr Stodola}.}
  \bibinfo{year}{2018}\natexlab{}.
\newblock \showarticletitle{Using Metaheuristics on the Multi-Depot Vehicle
  Routing Problem with Modified Optimization Criterion}.
\newblock \bibinfo{journal}{{\em Algorithms\/}} \bibinfo{volume}{11},
  \bibinfo{number}{5} (\bibinfo{year}{2018}), \bibinfo{pages}{74}.
\newblock


\bibitem[\protect\citeauthoryear{St{\"u}tzle and Hoos}{St{\"u}tzle and
  Hoos}{2000}]%
        {stutzle:2000}
\bibfield{author}{\bibinfo{person}{Thomas St{\"u}tzle} {and}
  \bibinfo{person}{Holger~H Hoos}.} \bibinfo{year}{2000}\natexlab{}.
\newblock \showarticletitle{MAX--MIN ant system}.
\newblock \bibinfo{journal}{{\em Future generation computer systems\/}}
  \bibinfo{volume}{16}, \bibinfo{number}{8} (\bibinfo{year}{2000}),
  \bibinfo{pages}{889--914}.
\newblock


\bibitem[\protect\citeauthoryear{Surekha and Sumathi}{Surekha and
  Sumathi}{2011}]%
        {surekha:2011}
\bibfield{author}{\bibinfo{person}{P Surekha} {and} \bibinfo{person}{S
  Sumathi}.} \bibinfo{year}{2011}\natexlab{}.
\newblock \showarticletitle{Solution to multi-depot vehicle routing problem
  using genetic algorithms}.
\newblock \bibinfo{journal}{{\em World Applied Programming\/}}
  \bibinfo{volume}{1}, \bibinfo{number}{3} (\bibinfo{year}{2011}),
  \bibinfo{pages}{118--131}.
\newblock


\bibitem[\protect\citeauthoryear{Thangiah and Salhi}{Thangiah and
  Salhi}{2001}]%
        {thangiah:2001}
\bibfield{author}{\bibinfo{person}{Sam~R Thangiah} {and} \bibinfo{person}{Said
  Salhi}.} \bibinfo{year}{2001}\natexlab{}.
\newblock \showarticletitle{Genetic clustering: an adaptive heuristic for the
  multidepot vehicle routing problem}.
\newblock \bibinfo{journal}{{\em Applied Artificial Intelligence\/}}
  \bibinfo{volume}{15}, \bibinfo{number}{4} (\bibinfo{year}{2001}),
  \bibinfo{pages}{361--383}.
\newblock


\bibitem[\protect\citeauthoryear{Tillman}{Tillman}{1969}]%
        {tillman:1969}
\bibfield{author}{\bibinfo{person}{Frank~A Tillman}.}
  \bibinfo{year}{1969}\natexlab{}.
\newblock \showarticletitle{The multiple terminal delivery problem with
  probabilistic demands}.
\newblock \bibinfo{journal}{{\em Transportation Science\/}}
  \bibinfo{volume}{3}, \bibinfo{number}{3} (\bibinfo{year}{1969}),
  \bibinfo{pages}{192--204}.
\newblock


\bibitem[\protect\citeauthoryear{Tillman and Hering}{Tillman and
  Hering}{1971}]%
        {tillman:1971}
\bibfield{author}{\bibinfo{person}{Frank~A Tillman} {and}
  \bibinfo{person}{Robert~W Hering}.} \bibinfo{year}{1971}\natexlab{}.
\newblock \showarticletitle{A study of a look-ahead procedure for solving the
  multiterminal delivery problem}.
\newblock \bibinfo{journal}{{\em Transportation Research\/}}
  \bibinfo{volume}{5}, \bibinfo{number}{3} (\bibinfo{year}{1971}),
  \bibinfo{pages}{225--229}.
\newblock


\bibitem[\protect\citeauthoryear{Wenjing and Ye}{Wenjing and Ye}{2010}]%
        {wenjing:2010}
\bibfield{author}{\bibinfo{person}{Zhang Wenjing} {and}
  \bibinfo{person}{Jianzhong Ye}.} \bibinfo{year}{2010}\natexlab{}.
\newblock \showarticletitle{An improved particle swarm optimization for the
  multi-depot vehicle routing problem}. In \bibinfo{booktitle}{{\em 2010
  International Conference on E-Business and E-Government}}. IEEE,
  \bibinfo{pages}{3188--3192}.
\newblock


\bibitem[\protect\citeauthoryear{Wren and Holliday}{Wren and Holliday}{1972}]%
        {wren:1972}
\bibfield{author}{\bibinfo{person}{Anthony Wren} {and} \bibinfo{person}{Alan
  Holliday}.} \bibinfo{year}{1972}\natexlab{}.
\newblock \showarticletitle{Computer scheduling of vehicles from one or more
  depots to a number of delivery points}.
\newblock \bibinfo{journal}{{\em Journal of the Operational Research
  Society\/}} \bibinfo{volume}{23}, \bibinfo{number}{3} (\bibinfo{year}{1972}),
  \bibinfo{pages}{333--344}.
\newblock


\bibitem[\protect\citeauthoryear{Yalian}{Yalian}{2016}]%
        {yalian:2016}
\bibfield{author}{\bibinfo{person}{Tang Yalian}.}
  \bibinfo{year}{2016}\natexlab{}.
\newblock \showarticletitle{An improved ant colony optimization for multi-depot
  vehicle routing problem}.
\newblock \bibinfo{journal}{{\em Int. J. Eng. Tech\/}}  \bibinfo{volume}{8}
  (\bibinfo{year}{2016}), \bibinfo{pages}{385--388}.
\newblock


\bibitem[\protect\citeauthoryear{Yao, Hu, Zhang, and Tian}{Yao
  et~al\mbox{.}}{2014}]%
        {yao:2014}
\bibfield{author}{\bibinfo{person}{Baozhen Yao}, \bibinfo{person}{Ping Hu},
  \bibinfo{person}{Mingheng Zhang}, {and} \bibinfo{person}{Xiaomei Tian}.}
  \bibinfo{year}{2014}\natexlab{}.
\newblock \showarticletitle{Improved ant colony optimization for seafood
  product delivery routing problem}.
\newblock \bibinfo{journal}{{\em PROMET-Traffic\&Transportation\/}}
  \bibinfo{volume}{26}, \bibinfo{number}{1} (\bibinfo{year}{2014}),
  \bibinfo{pages}{1--10}.
\newblock


\bibitem[\protect\citeauthoryear{Yu, Yang, and Xie}{Yu et~al\mbox{.}}{2011}]%
        {yu:2011}
\bibfield{author}{\bibinfo{person}{Bin Yu}, \bibinfo{person}{ZZ Yang}, {and}
  \bibinfo{person}{JX Xie}.} \bibinfo{year}{2011}\natexlab{}.
\newblock \showarticletitle{A parallel improved ant colony optimization for
  multi-depot vehicle routing problem}.
\newblock \bibinfo{journal}{{\em Journal of the Operational Research
  Society\/}} \bibinfo{volume}{62}, \bibinfo{number}{1} (\bibinfo{year}{2011}),
  \bibinfo{pages}{183--188}.
\newblock


\end{thebibliography}

\end{document}